\documentclass[11pt,a4paper]{article}
\usepackage[hyperref]{acl2020}
\usepackage{times}
\usepackage{latexsym}
\usepackage[ruled,vlined,linesnumbered]{algorithm2e}
\usepackage{amsmath}
\usepackage{graphicx}  

\usepackage{microtype}

\aclfinalcopy 
\def\aclpaperid{67} 

\title{Using Random Perturbations to Mitigate Adversarial \\Attacks on Sentiment Analysis Models}

\author{Abigail Swenor \and Jugal Kalita\\
  University of Colorado Colorado Springs\\
  1420 Austin Bluffs Pkwy \\
  Colorado Springs, CO 80918 \\
  \texttt{aswenor,jkalita@uccs.edu} }

\date{}

\begin{document}
\maketitle
\begin{abstract}
 Attacks on deep learning models are often difficult to identify and therefore are difficult to protect against. This problem is exacerbated by the use of public datasets that typically are not manually inspected before use. In this paper, we offer a solution to this vulnerability by using, during testing, random perturbations such as spelling correction if necessary, substitution by random synonym, or simply dropping the word. These perturbations are applied to random words in random sentences to defend NLP models against adversarial attacks. Our Random Perturbations Defense and Increased Randomness Defense methods are successful in returning attacked models to similar accuracy of models before attacks. The original accuracy of the model used in this work is 80\% for sentiment classification. After undergoing attacks, the accuracy drops to accuracy between 0\% and 44\%. After applying our defense methods, the accuracy of the model is returned to the original accuracy within statistical significance.
\end{abstract}

\section{Introduction}

\noindent Deep learning models have excelled in solving difficult problems in machine learning, including Natural Language Processing (NLP) tasks like text classification \citep{zhang2015character,kim2014convolutional} and language understanding \citep{devlin2019bert}. However, research has discovered that inputs can be modified to cause trained deep learning models to produce incorrect results and predictions \citep{szegedy2014intriguing}. Models in computer vision are vulnerable to these attacks \citep{goodfellow2015explaining}, and studies have found that models in the NLP domain are also vulnerable \citep{kuleshov2018adversarial,gao2018black,garg2020bae}. One use of these adversarial attacks is to test and verify the robustness of NLP models. 
\\\indent With the potential for adversarial attacks, there comes the need for prevention and protection. There are three main categories of defense methods: identification, reconstruction, and prevention \citep{goldblum2020data}. Identification methods rely on detecting either poisoned data or the poisoned model \citep{ijcai2019-647}. While reconstruction methods actively work to repair the model after training \citep{10.1145/3394171.3413546}, prevention methods rely on input preprocessing, majority voting, and other techniques to mitigate adversarial attacks \citep{goldblum2020data,alshemali2020generalization}. Although most NLP adversarial attacks are easily detectable, some new forms of adversarial attacks have become more difficult to detect like concealed data poisoning attacks \citep{wallace2021concealed} and backdoor attacks \citep{chen2020badnl}. The use of these concealed and hard-to-detect attacks has revealed new vulnerabilities in NLP models. Considering the increasing difficulty in detecting attacks, a more prudent approach would be to work on neutralizing the effect of potential attacks rather than solely relying on detection. Here we offer a novel and highly effective defense solution that preprocesses inputs by random perturbations to mitigate potential hard-to-detect attacks. 

\section{Related Work}
\indent The work in this paper relates to the attack on NLP models using the TextAttack library \citep{morris2020textattack}, the current state-of-the-art defense methods for NLP models, and using randomness against adversarial attacks. 
\\\indent The TextAttack library and the associated GitHub repository \citep{morris2020textattack} represent current efforts to centralize attack and data augmentation methods for the NLP community. The library supports attack creation through the use of four components: a goal function, a search method, a transformation, and constraints. An attack method uses these components to perturb the input to fulfill the given goal function while complying with the constraints and the search method finds transformations that produce adversarial examples. The library contains a total of 16 attack model recipes based on literature. The work reported in this paper pertains to the 14 ready-to-use classification attack recipes from the TextAttack library. We believe that successful defense against such attacks will provide guidelines for the general defense of deep learning NLP classification models.
\\\indent There are many methods to defend NLP models against adversarial attacks, including input preprocessing. Input preprocessing defenses require inserting a step between the input and the given model that aims to mitigate any potential attacks. \citet{alshemali2020generalization} use an input preprocessing defense that employs synonym set averages and majority voting to mitigate synonym substitution attacks. Their method is deployed before the input is run through a trained model. Another defense against synonym substitution attacks, Random Substitution Encoding (RSE) encodes randomly selected synonyms to train a robust deep neural network \citep{10.1007/978-3-030-55393-7_28}. The RSE defense occurs between the input and the embedding layer. 
\\\indent Randomness has been deployed in computer vision defense methods against adversarial attacks. \citet{levine2020robustness} use random ablations to defend against adversarial attacks on computer vision classification models. Their defense is based on a random-smoothing technique that creates certifiably robust classification. \citeauthor{levine2020robustness} defend against sparse adversarial attacks that perturb a small number of features in the input images. They found their random ablation defense method to produce certifiably robust results on the MNIST, CIFAR-10, and ImageNet datasets. 

\section{Input Perturbation Approach \& Adversarial Defense}
\noindent The use and availability of successful adversarial attack methods reveal the need for defense methods that do not rely on detection and leverage intuitions gathered from popular attack methods to protect NLP models. In particular, we present a simple but highly effective defense against attacks on deep learning models that perform sentiment analysis. 
\\\indent The approach taken is based on certain assumptions about the sentiment analysis task. Given a short piece of text, we believe that a human does not need to necessarily analyze every sentence carefully to get a grasp on the sentiment. Our hypothesis is that humans can ascertain the expressed sentiment in a text by paying attention to a few key sentences while ignoring or skimming over the others. This thought experiment led us to make intermediate classifications on individual sentences of a review in the IMDB dataset and then combining the results for a collective final decision.
\\\indent This process was refined further by considering how attackers actually perturb data. Usually, they select a small number of characters or tokens within the original data to perturb. To mitigate those perturbations, we choose to perform our own random perturbations. Because the attacking perturbations could occur anywhere within the original data, and we do not necessarily know where they are, it is prudent to randomly select tokens for us to perturb. This randomization has the potential to negate the effect the attacking perturbations have on the overall sentiment analysis. 
\\\indent We wish to highlight the importance of randomness in our approach and in possible future approaches for defenses against adversarial attacks. Positive impact of randomness in classification tasks with featured datasets can be found in work using Random Forests \citep{breiman2001random}. Random Forests have been useful in many domains to make predictions, including disease prediction \citep{lebedev2014random,corradi2018prediction,paul2017feature,khalilia2011predicting} and stock market price prediction \citep{khaidem2016predicting,ballings2015evaluating,nti2019random}. The use of randomness has made these methods of prediction robust and useful. We have chosen to harness the capability of randomness in defense of adversarial attacks in NLP. We demonstrate that the impact randomness has on our defense method is highly positive and its use in defense against adversarial attacks of neural networks should be explored further. We present two algorithms below---first with two levels of randomness, and the second with three.

\subsection{Random Perturbations Defense}
\begin{figure}[h]
\centering
\includegraphics[scale=.2]{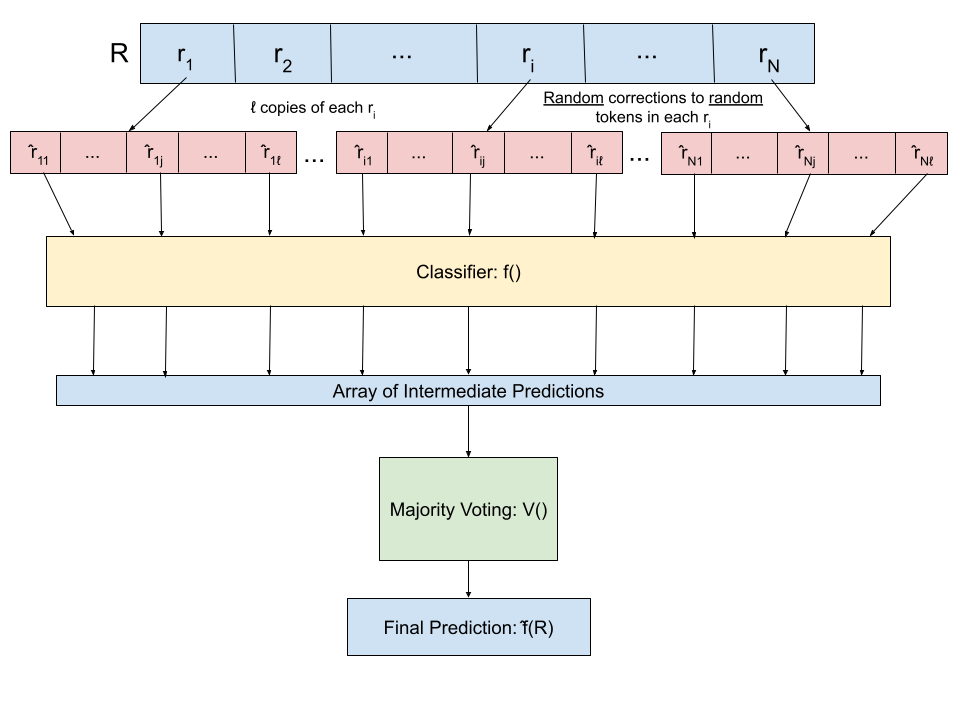}
\caption{Visual representation of Algorithm \ref{algorithm:algo}.}
\label{fig:AlgoImage}
\end{figure}

\noindent Our algorithm is based on random processes: the randomization of perturbations of the sentences of a review $R$ followed by majority voting to decide the final prediction for sentiment analysis. We consider each review $R$ to be represented as a set $R$ = $\{r_1, r_2,..., r_i,..., r_N\}$ of sentences $r_i$. Once $R$ is broken down into its sentences (Line 1 of Algorithm \ref{algorithm:algo}), we create $l$ replicates of sentence $r_i$: $\{\hat{r}_{i1},...,\hat{r}_{ij},...,\hat{r}_{il}\}$. Each replicate $\hat{r}_{ij}$ has $k$ number of perturbations made to it. Each perturbation is determined randomly (Lines 4-7). 
\\\indent In Line 5, a random token $t$ where $t \in \hat{r}_{ij}$ is selected, and in Line 6, a random perturbation is performed on $t$. This random perturbation could be a spellcheck with correction if necessary, a synonym substitution, or dropping the word. These perturbations were selected as they are likely to be the same operations an attacker performs, and they may potentially even counter the effect of a large portion of perturbations in attacked data. A spellcheck is performed using SpellChecker which is based in Pure Python Spell Checking. If a spellcheck is performed on a token without spelling error, then the token will not be changed. The synonym substitution is also performed in a random manner. A synonym set for token $t$ is found using the WordNet synsets \citep{wordnet}. Once a synonym set is found, it is processed to remove any duplicate synonyms or copies of token $t$. Once the synonym set is processed, a random synonym from the set is chosen to replace token $t$ in $\hat{r}_{ij}$. A drop word is when the randomly selected token $t$ is removed from the replicate altogether and replaced with a space. Conceptually speaking, the random perturbations may be chosen from an extended set of allowed changes.
\\\indent Once $l$ replicates have been created for the given sentence $r_i$ and perturbations made to tokens, they are put together to create replicate review set $\hat{R}$ (Line 8). Then, in Line 9, each $\hat{r}_{ij} \in \hat{R}$ is classified individually as $f(\hat{r}_{ij})$ using classifier $f()$. After each replicate has been classified, we perform majority voting with function $V()$. We call the final prediction that this majority voting results in as $\hat{f}(R)$. This function can be thought of as follows (Line 12): $$\hat{f}(R) = V(\{f(\hat{r}_{ij})\ |\ \hat{r}_{ij} \in \hat{R}\}).$$
The goal is to maximize the probability that $\hat{f}(R) = f(R)$ where $f(R)$ is the classification of the original review $R$. In this paper, this maximization is done through tuning of the parameters $l$ and $k$. The certainty $T$ for $\hat{f}(R)$ is also determined for each calculation of $\hat{f}(R)$. The certainty represents how sure the algorithm is of the final prediction it has made. In general, the certainty $T$ is determined as follows (Lines 13-17): $$T = count(f(\hat{r}_{ij}) == \hat{f}(R))\ /\ N*l.$$
The full visual representation of this algorithm can be seen in Algorithm \ref{algorithm:algo} and in Figure \ref{fig:AlgoImage}.

\begin{algorithm}[h]
\SetAlgoLined
\SetKwInOut{Input}{Input}
\SetKwInOut{Output}{Parameters}
\KwResult{$\hat{f}(R)$, the classification of $R$ after defense}
\Input{Review $R = \{r_1, r_2,...,r_N\}$ where $r_i$ is a sentence}
\Output{$l$ = number of copies made of each $r$, $k$ = number of corrections made per $r_i$, $C = \{c_1, c_2,..., c_k\}$, set of corrections}
$\hat{R} = \emptyset$\\
\For{$r_i\ \epsilon\ R$}{
    \For{j = 1 to l}{
        $\hat{r}_{ij} = r_i$\\
        \For{k}{
            Select random token $t$ where $t \ \epsilon \ \hat{r}_{ij}$
            \\Perform random correction $c\ \epsilon\ C$ to $t$
            }
        Append $\hat{r}_{ij}$\ to\ $\hat{R}$
        \\Classify: $f(r_{ij})$
        }
}
$\hat{f}(R) = V(\{f(\hat{r}_{ij}) \ |\ \hat{r}_{ij} \epsilon \hat{R}\})$, $V()$ is a voting function
\\\uIf{$f(\hat{R}) == negative$}{
    $T = count(f(\hat{r}_{ij}) == negative)\ /\ N*l$
    }
\Else{
    $T = count(f(\hat{r}_{ij} == positive)\ /\ N*l$
    }
\caption{Random Perturbation Defense}
\label{algorithm:algo}
\end{algorithm}

\subsection{Increasing Randomness}
\begin{figure}[h]
\centering
\includegraphics[scale=.2]{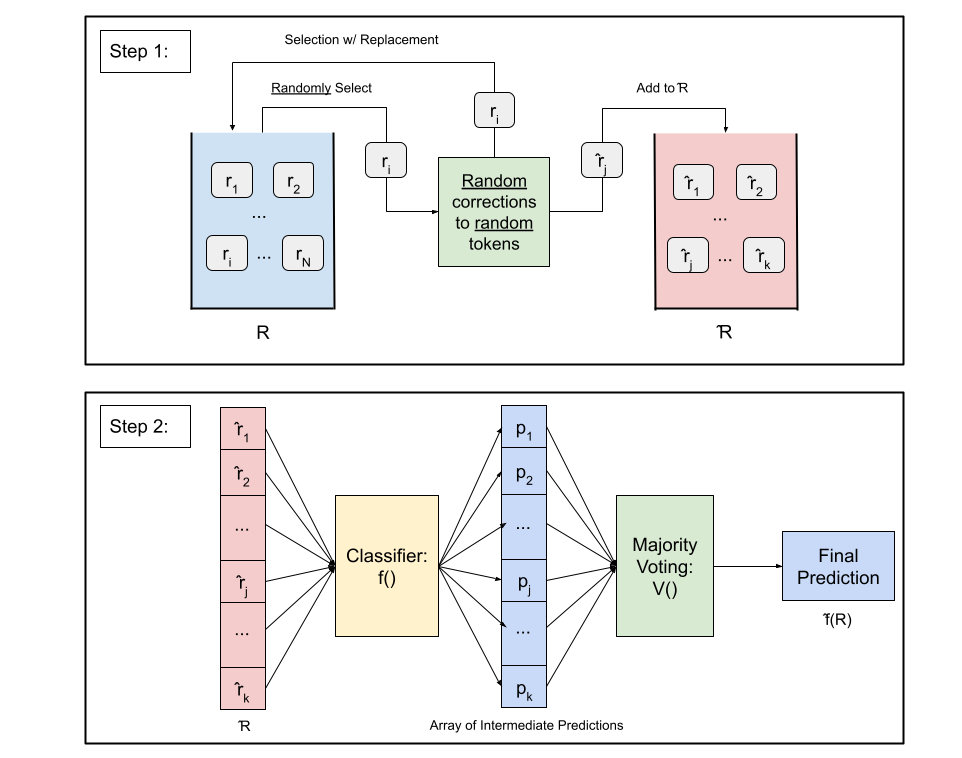}
\caption{Visual representation of Algorithm \ref{algorithm:MoreRandAlgo} that includes more randomness.}
\label{fig:MoreRandImage}
\end{figure}

\noindent Our first algorithm represented in Algorithm \ref{algorithm:algo} and in Figure \ref{fig:AlgoImage} shows randomness in two key points in the decision making process for making the perturbations. This is the main source of randomness for our first algorithm. In our next algorithm, we introduce more randomness into our ideas from our original algorithm to create a modified algorithm. This more random algorithm is visually represented in Figure \ref{fig:MoreRandImage} and presented in Algorithm \ref{algorithm:MoreRandAlgo}. This new defense method adds a third random process before making random corrections to a sentence. Randomly chosen $r_i$ from $R$ are randomly corrected to create replicate $\hat{r}_{j}$ which is placed in $\hat{R}$ (Lines 2-6). The original sentence $r_i$ is placed back into $R$ and a new sentence is randomly selected; this is random selection with replacement. This process of random selection is repeated until there is a total of $k$ replicates $\hat{r}_{j}$ in $\hat{R}$. This algorithm follows the spirit of Random Forests more closely than the first algorithm.
\\\indent In Line 2, we randomly select a sentence $r_i$ from $R$. This is one of the main differences between Algorithm \ref{algorithm:algo} for Random Perturbations Defense and Algorithm \ref{algorithm:MoreRandAlgo} for Increased Randomness Defense. That extra random element allows for more randomization in the corrections we make to create replicates $\hat{r}_j$. In Lines 3 and 4, the process is practically identical to Lines 5 and 6 in Algorithm \ref{algorithm:algo}. The only difference is that only one random correction is being made to get the final replicate $\hat{r}_j$ for Increased Randomness Defense, while Random Perturbations Defense makes $k$ random corrections to get the final replicate $\hat{r}_{ij}$.

\begin{algorithm}[h]
\SetAlgoLined
\SetKwInOut{Input}{Input}
\SetKwInOut{Output}{Parameters}
\KwResult{$\hat{f}(R)$, the classification of $R$ after defense}
\Input{Review $R = \{r_1, r_2,...,r_N\}$ where $r_i$ is a sentence}
\Output{$k$ = number of replicates $\hat{r}_j$ made for $\hat{R}$, $C = \{c_1, c_2,..., c_k\}$, set of corrections}
$\hat{R} = \emptyset$, $P = []$\\
\For{j = 1 to k}{
    Randomly select $r_i \in R$
    \\Select random token $t$ where $t \in r_i$
    \\Perform random correction $c\ \epsilon\ C$ to $t$ to get $\hat{r}_j$
    \\Append $\hat{r}_j$ to $\hat{R}$
}
\For{j = 1 to k}{
    Classify: $f(\hat{r}_j)$
    \\Append results to predictions array $P$
}
$\hat{f}(R) = V(P)$, $V()$ is a voting function

\uIf{$f(\hat{R}) == negative$}{
    $T = count(f(\hat{r}_{ij}) == negative)\ /\ N*l$
    }
\Else{
    $T = count(f(\hat{r}_{ij} == positive)\ /\ N*l$
    }
\caption{Increased Randomness Defense}
\label{algorithm:MoreRandAlgo}
\end{algorithm}

\subsection{Overcoming the Attacks}
\noindent We define an attack as making random perturbations to an input, specifically for this work, a review $R$. We assume a uniform distribution for randomness. We interpret these random changes to occur throughout each review $R$ with probability $\frac{1}{W}$ or $\frac{1}{N*m}$, where $W$ is the number of words in $R$, $N$ is the number of sentences in $R$, and $m$ is the average length of each sentence in $R$. We refer to this probability that an attack makes changes to the review text as $P_{attack}$ where $a$ is the total number of perturbations made by the attack: $$P_{attack} = \frac{a}{W} = \frac{a}{N*m}.$$ If each random perturbation performed by the attack has a probability of $\frac{1}{N*m}$, then our defense method needs to overcome that probability to overcome the attack.
\\\indent Our two defense methods, Random Perturbations Defense and Increased Randomness Defense, both offer ways to overcome the attack, i.e., undo the attack change, with a probability greater than $\frac{a}{N*m}$. 
\newtheorem{theorem}{Proposition}
\begin{theorem}
Random Perturbations Defense overcomes an attack that makes a small number of random perturbations to a review document by having a probability greater than the attack probability $P_{attack}$.
\end{theorem}
\indent Our Random Perturbations Defense picks a random token $t$ from each sentence $r_i \in R$ and repeats $k$ times to get a final replicate $\hat{r}_{ij}$. This gives an initial probability that the defense picks a certain token from the text, or $P_{RPD}$, to be: $$P_{RPD} = \frac{N*l*m!}{k!(m-k)!}.$$ We find this probability from choosing $k$ tokens from $r_i$ with length $m$ which breaks down to a binomial coefficient $\binom{m}{k} = \frac{m!}{k!(n-k)!}$. This is then repeated $l$ times for each sentence in $R$ which equates to that initial probability being multiplied by $l$ and $N$. After doing some rearranging of the probabilities, we can see that for certain values of $l$ and $k$ where $k < m$: \small \[P_{RPD} = \frac{N^2 m^2 l(m-1)(m-2)...(m-k+1)}{k!} > a.\] \normalsize
$P_{RPD}$ now is the total probability that the defense makes random changes to $lN$ tokens. We know that $W = N*m$, that $a =< W$ for the attack methods we are testing against, and that $k$ should be selected so that $k << W$. This means that we know $W^2 > a$, $W^2 > k!$, and $l(m-1)(m-2)...(m-k+1) > 0$ for the selected attack methods, which gives us the necessary conditions to assert that $P_{RPD} > P{attack}$. Therefore, our Random Perturbations Defense will overcome the $P_{attack}$ and should overcome the given attack method as stated in Proposition 1. 

\begin{theorem}
Increased Randomness Defense overcomes an attack that makes a small number of random perturbations to a review document by having a probability greater than the attack probability $P_{attack}$.
\end{theorem}
\indent Our Increased Randomness Defense first chooses a random sentence $r_i$ which is selected with probability $\frac{1}{N}$. Next, we choose a random word within that sentence which is selected with probability $\frac{1}{m}$. This gives us a probability for changes as follows: $$P_{IRD} = \frac{1}{N} * \frac{1}{m} = \frac{1}{N*m}.$$ We can see that $P_{IRD} * a = P_{attack}$. We need to overcome the attack probability and we do this in two ways: we either find the attack perturbation by chance and reverse it, or we counterbalance the attack perturbation with enough replicates $\hat{r}_j$. With each replicate $\hat{r}_j$ created, we increase our probability $P_{IRD}$ so that our final probability for our Increased Randomness Defense is as follows: $$P_{IRD} = \frac{k}{N*m}.$$ As long as our selected parameter value for $k$ is greater than the number of perturbation changes made by the attack method $a$, then $P_{IRD} > P_{attack}$ and our Increased Randomness Defense method will overcome the given attack method as stated in Proposition 2. 

\section{Experiments \& Results}
\subsection{Dataset \& Models}
\noindent We used the IMDB dataset \citep{maas-EtAl:2011:ACL-HLT2011} for our experiments. Each attack was used to perturb 100 reviews from the dataset. The 100 reviews were selected randomly from the dataset with a mix of positive and negative sentiments. Note that the Kuleshov attack data \citep{kuleshov2018adversarial} only had 77 reviews.
\\\indent The models used in this research are from the TextAttack \citep{morris2020textattack} and HuggingFace \citep{wolf-etal-2020-transformers} libraries. These libraries offer many different models to use for both attacked data generation and general NLP tasks. For this research, we used the \emph{bert-base-uncased-imdb} model that resides in both the TextAttack and HuggingFace libraries. This model was fine-tuned and trained with a cross-entropy loss function. This model was used with the API functions of the TextAttack library to create the attacked reviews from each of the attacks we used. We chose this model because BERT models are useful in many NLP tasks and this model specifically was fine-tuned for text classification and was trained on the dataset we wanted to use for these experiments. 
\\\indent The HuggingFace library was also used in the sentiment-analysis classification of the attacked data and the defense method. We used the HuggingFace transformer pipeline for sentiment-analysis to test our defense method. This pipeline returns either ``negative" or ``positive" to classify the sentiment of the input text and a score for that prediction \citep{wolf-etal-2020-transformers}. This pipeline was used to classify each replicate $\hat{r}_{ij}$ in our algorithm and is represented as the function $f()$.
\subsection{Experiments}
\begin{figure}[ht]
\centering
\includegraphics[scale=.28]{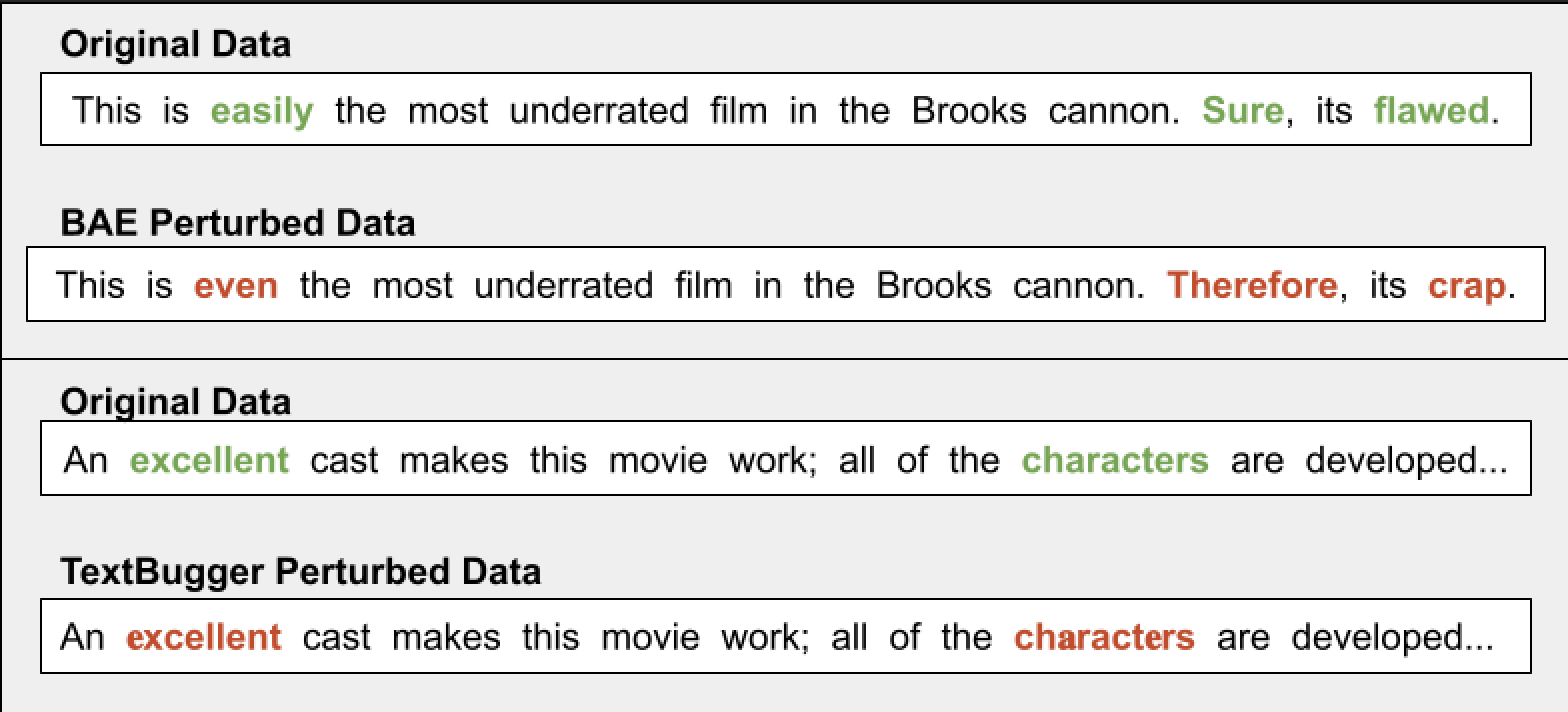}
\caption{Example of what original data looks like and how the BAE \cite{garg2020bae} and TextBugger \cite{li2019textbugger} attack methods perturb data. The BAE attack method uses semantic similarity, while the Textbugger attack method uses visual similarity. }
\label{fig:perturbedData}
\end{figure}

\noindent The attacks from the TextAttack library were used to generate attack data. Attack data was created from 7 different models from the library: BERT-based Adversarial Examples (BAE) \citep{garg2020bae}, DeepWordBug \citep{gao2018black}, FasterGeneticAlgorithm \citep{jia2019certified}, Kuleshov \citep{kuleshov2018adversarial}, Probability Weighted Word Saliency (PWWS) \citep{ren2019generating}, TextBugger \citep{li2019textbugger}, and TextFooler \citep{jin2020bert}  \citep{morris2020textattack}. Each of these attacks were used to create 100 perturbed sentences from the IMDB dataset \citep{maas-EtAl:2011:ACL-HLT2011}. These attacks were chosen from the 14 classification model attacks because they represent different kinds of attack methods, including misspelling, synonym substitution, and antonym substitution. 
\\\indent Each attack method used for our experiments has a slightly different approach to perturbing the input data. Each perturbation method is unique and follows a specific distinct pattern and examples of these can be found in Figure \ref{fig:perturbedData}. The BAE attack determines the most important token in the input and replaces that token with the most similar replacement using a Universal Sentence Encoder. This helps the perturbed data remain semantically similar to the original input \citep{garg2020bae}. The DeepWordBug attack identifies the most important tokens in the input and performs character-level perturbations on the highest-ranked tokens while minimizing edit distance to create a change in the original classification \citep{gao2018black}. The FasterGeneticAlgorithm perturbs every token in a given input while maintaining the original sentiment. It chooses each perturbation carefully to create the most effective adversarial example \citep{jia2019certified}. The Kuleshov attack is a synonym substitution attack that replaces 10\% - 30\% of the tokens in the input with synonyms that do not change the meaning of the input \citep{kuleshov2018adversarial}. 
\\\indent The PWWS attack determines the word saliency score of each token and performs synonym substitutions based on the word saliency score and the maximum effectiveness of each substitution \citep{ren2019generating}. The TextBugger attack determines the important sentences from the input first. It then determines the important words in those sentences and generates 5 possible ``bugs" through different perturbation methods: insert, swap, delete, sub-c (visual similarity substitution), sub-w (semantic similarity substitution). The attack will implement whichever of these 5 generated bugs is the most effective in changing the original prediction \citep{li2019textbugger}. Finally, the TextFooler attack determines the most important tokens in the input using synonym extraction, part-of-speech checking, and semantic similarity checking. If there are multiple canididates to substitute with, the most semantically similar substitution will be chosen and will replace the original token in the input \citep{jin2020bert}. 

\indent After each attack had corresponding attack data, the TextAttack functions gave the results for the success of the attack. The accuracy of the sentiment-analysis task under attack, without the defense method, is reported in the first column in Table \ref{tab:results}. Each attack caused a large decrease in the accuracy of the model. The model began with an average accuracy of 80\% for the IMDB dataset. Once the attack data was created and the accuracy under attack was reported, the attack data was run through our Random Perturbations and Increased Randomness defense methods. All of the experiments were run on Google Colaboratory using TPUs and the Natural Language Toolkit \citep{Loper02nltk:the}. 

\subsection{Results}
\noindent We began by testing on the HuggingFace sentiment analysis pipeline with the original IMDB dataset. This gave an original accuracy of 80\%. This percentage represents the goal for our defense method accuracy as we aim to return the model to its original accuracy, or higher. The accuracy under each attack is listed in Table \ref{tab:results} in the first column. These percentages show how effective each attack is at causing misclassification for the sentiment analysis task. The attacks range in effectiveness with PWWS \citep{ren2019generating} and Kuleshov \citep{kuleshov2018adversarial} with the most successful attacks at 0\% accuracy under attack and FasterGeneticAlgorithm \citep{jia2019certified} with the least successful attack at 44\% accuracy under attack, which is still almost a 40\% drop in accuracy. 
\begin{table}[h]
    \centering
    \resizebox{\columnwidth}{!}{
    \begin{tabular}{|c|c|c|}
    \hline \textbf{Attack} & \textbf{w/o Defense} & \textbf{w/ Defense}  \\
    \hline\hline
    BAE & 33\% & 80.80\%$\pm$1.47 \\
    \hline
    DeepWordBug & 34\% & 76.60\%$\pm$1.85 \\
    \hline
    FasterGeneticAlgo & 44\% & 82.20\%$\pm$1.72 \\
    \hline
    Kuleshov* & 0\% & 60.00\%$\pm$2.24 \\
    \hline
    PWWS & 0\% & 81.80\%$\pm$1.17 \\
    \hline
    TextBugger & 6\% & 79.20\%$\pm$2.32 \\
    \hline
    TextFooler & 1\% & 83.20\%$\pm$2.48 \\
    \hline
    \end{tabular}}
    \caption{Accuracy for each of the attack methods under attack, and under attack with the defense method from Algorithm \ref{algorithm:algo} deployed with $l=7$ and $k=5$. The accuracy prior to attack is 80\%.}
    \label{tab:results}
\end{table}

\subsubsection{Random Perturbations Defense} For the Random Perturbations Defense to be successful, it is necessary to obtain values of the two parameters, $l$ and $k$. Each attack was tested against our Random Perturbations Defense 5 times. The accuracy was averaged for all 5 tests and the standard deviation was calculated for the given mean. The mean accuracy with standard deviation is presented for each attack in the second column of Table \ref{tab:results}. The results presented are for $l = 7$ and $k = 5$. These parameters were chosen after testing found greater values of $l$ and $k$ resulted in a longer run time and too many changes made to the original input; with lower values for $l$ and $k$, the model had lower accuracy and not enough perturbations to outweigh any potential adversarial attacks. The values behind this logic can be seen in Table \ref{tab:landk}.
\begin{table}[h]
    \centering
    \begin{tabular}{|c|c|c|c|}
    \hline \textbf{Attack} & \textbf{l} & \textbf{k} & \textbf{Accuracy w/ Defense}\\
    \hline\hline
    BAE & 5 & 2 & 55\% \\
    \hline
    BAE & 10 & 5 & 50\% \\
    \hline
    BAE & 7 & 5 & 79\% \\
    \hline
    \end{tabular}
    \caption{This table explains values of $l$ and $k$}
    \label{tab:landk}
\end{table}

\indent The defense method was able to return the model to original accuracy within statistical significance while under attack for most of the attacks with the exception of the Kuleshov method \citep{kuleshov2018adversarial}. The accuracy for the other attacks all were returned to the original accuracy ranging from 76.00\% to 83.20\% accuracy with the Random Perturbations defense deployed. This shows that our defense method is successful at mitigating most potential adversarial attacks on sentiment classification models. Our defense method was able to increase the accuracy of model while under attack for the FasterGeneticAlgorithm, PWWS, and TextFooler. These three attack methods with our defense achieved accuracy that was higher than the original accuracy with statistical significance.

\subsubsection{Increased Randomness Defense} The Increased Randomness Defense was also tested on all seven of the attacks. Each attack was tested against this defense 5 times. The results for these experiments can be seen in Table \ref{tab:MoreRandRes}. There were tests done to determine what the proper value for $k$ should be. These tests were performed on the BAE \citep{garg2020bae} attack and the results can be found in Table \ref{tab:MoreRandk}. These tests revealed that 40-45 replicates $\hat{r}_j$ was ideal for each $\hat{R}$ with $k = 41$ being the final value used for the tests on each attack. This defense method was more efficient to use. 

\begin{table}[h]
    \centering
    \begin{tabular}{|c|c|c|}
    \hline \textbf{Attack} & \textbf{k} & \textbf{Accuracy w/ Defense} \\
    \hline\hline
    BAE & 10 & 67\% \\
    \hline
    BAE & 20 & 76\% \\
    \hline
    BAE & 25 & 72\% \\
    \hline
    BAE & 30 & 76\% \\
    \hline
    BAE & 35 & 74\% \\
    \hline
    BAE & 40 & 82\% \\
    \hline
    BAE & 45 & 74\% \\
    \hline
    BAE & 41 & 77\% \\
    \hline
    \end{tabular}
    \caption{This table shows the results for the tests for different values of $k$ for the increased randomness experiments.}
    \label{tab:MoreRandk}
\end{table}

\begin{figure}[h]
\centering
\includegraphics[scale=.29]{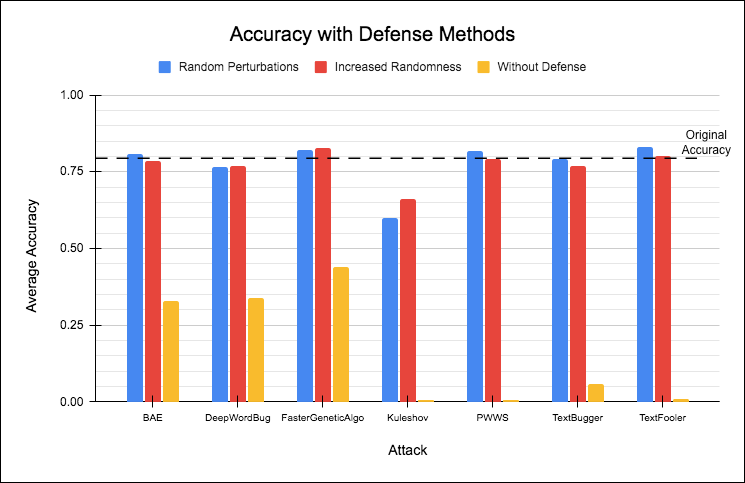}
\caption{Comparing the average accuracy of the Random Perturbations Defense and the Increased Randomness Defense methods to the under attack accuracy without defense on the seven attacks.}
\label{fig:Graph}
\end{figure}

The runtime and the resources used for this method were lower than the original random perturbations defense method with the runtime for the Random Perturbations Defense being nearly 4 times longer than this increased random method. A comparison of the two defense methods on the seven attacks tested can be seen in Figure \ref{fig:Graph}. This defense was successful in returning the model to the original accuracy, within statistical significance, for most of the attacks with the exception of the Kuleshov attack \citep{kuleshov2018adversarial}. A t-test was performed to determine the statistical significance of the difference in the defense method accuracy to the original accuracy. 

\begin{table}[h]
    \centering
    \resizebox{\columnwidth}{!}{
    \begin{tabular}{|c|c|c|}
    \hline \textbf{Attack} & \textbf{w/o Defense} & \textbf{w/ Defense} \\
    \hline\hline
    BAE & 33\% & 78.40\%$\pm$3.14 \\
    \hline
    DeepWordBug & 34\% & 76.80\%$\pm$2.64 \\
    \hline
    FasterGeneticAlgo & 44\% & 82.80\%$\pm$2.48 \\
    \hline
    Kuleshov* & 0\% & 66.23\%$\pm$4.65 \\
    \hline
    PWWS & 0\% & 79.20\%$\pm$1.72 \\
    \hline
    TextBugger & 6\% & 77.00\%$\pm$2.97 \\
    \hline
    TextFooler & 1\% & 80.20\%$\pm$2.48 \\
    \hline
    \end{tabular}}
    \caption{Accuracy for increased randomness defense from Algorithm \ref{algorithm:MoreRandAlgo} against each attack method with $k = 41$. The accuracy prior to attack is 80\%.}
    \label{tab:MoreRandRes}
\end{table}

\subsection{Comparison to Recent Defense Methods}
Our defense methods are comparable to some recent defense methods created for text classification. Our defense method returns the model to the original accuracy within statistical significance. This is comparable to the work done by \citet{zhou2021defense} in their Dirichlet Neighborhood Ensemble (DNE) defense method. They were able to bring the model within 10\% of the original accuracy for CNN, LSTM, and BOW models for the IMDB dataset. However, their work is only applicable to synonym-substitution based attacks. Since our defense methods apply equally well to seven attacks, it is general and can be applied without determining the exact type of attack (assuming it is one of the seven). 
\\\indent Another recent defense method, Synonym Encoding Method (SEM), was tested on synonym-substitution attacks on Word-CNN, LSTM, Bi-LSTM and BERT models \citep{wangnatural}. This defense method was most successful on the BERT model and was able to return to the original accuracy within 3\% for the IMDB dataset. Our work is comparable to both DNE and SEM which represent recent work in defending NLP models against adversarial attacks and more specifically synonym-substitution based attacks. 
\\\indent WordDP is another recent defense method for adversarial attacks against NLP models \citep{wang-etal-2021-certified}. This defense method used Differential Privacy (DP) to create certified robust text classification models against word substitution adversarial attacks. They tested their defense on the IMDB and found that their WordDP method was successful at raising the accuracy within 3\% of the original clean model. This method outperformed other defense methods including DNE. This is similar to our defense method, but they do not include whether these results are statistically significant. 
\\\indent We also compare our defense methods, RPD and IRD, against these recent defense methods on cost and efficiency. Our RPD and IRD methods have comparable time complexity of $O(cn)$, where $c$ is the time it takes for classification and $n$ is the number of reviews. Each method has a similar constant that represents the number of perturbations and replicates made. We cannot directly compare the time complexity of our defense methods with the SEM, DNE, and WordDP methods. These recent defense methods require specialized training and/or encodings. Our RPD and IRD methods do not require specialized training or encodings, so they cannot be directly compared on time complexity. This means that the comparison between our methods and recent defense methods comes in the form of specialized training vs. input preprocessing. Training and developing new encodings tends to be more time consuming and expensive than input preprocessing methods that can occur during the testing phases. 

\section{Conclusion}
The work in this paper details a successful defense method against adversarial attacks generated from the TextAttack library. These attack methods use multiple different perturbation approaches to change the predictions made by NLP models. Our Random Perturbations Defense was successful in mitigating 6 different attack methods. This defense method returned the attacked models to their original accuracy within statistical significance. Our second method, Increased Randomness Defense, used more randomization to create an equally successful defense method that was 4 times more efficient than our Random Perturbations Defense. Overall, our defense methods are effective in mitigating a range of NLP adversarial attacks, presenting evidence for the effectiveness of randomness in NLP defense methods. The work done here opens up further study into the use of randomness in defense of adversarial attacks for NLP models including the use of these defense methods for multi-class classification. This work also encourages a further mathematical and theoretical explanation to the benefits of randomness in defense of NLP models.

\section*{Acknowledgement}
The work reported in this paper is supported by the National Science Foundation under Grant No. 2050919. Any opinions, findings and conclusions or recommendations expressed in this work are those of the author(s) and do not necessarily reflect the views of the National Science Foundation.

\bibliography{SwenorICON21}
\bibliographystyle{acl_natbib}
\end{document}